\documentclass[letterpaper, 10 pt, conference]{ieeeconf}  %

\usepackage[letterpaper, left=.7in, right=.8in, bottom=.76in, top=.75in]{geometry}
\IEEEoverridecommandlockouts                              %

\usepackage{graphics} %
\usepackage{epsfig} %
\usepackage{epstopdf}
\usepackage{mathptmx} %
\usepackage{times} %
\usepackage{amsmath} %
\usepackage{amssymb}  %
\usepackage{color}
\usepackage{tabularx,booktabs}
\usepackage{siunitx}
\usepackage[font=small]{caption}
\usepackage{subcaption}
\usepackage{hyperref}
\usepackage{float}
\usepackage{dblfloatfix}
\usepackage{nidanfloat}
\usepackage{multirow}
\usepackage{comment}
\usepackage{tikz}
\usetikzlibrary{positioning,chains}
\title{\LARGE \bf Fundamental Challenges in Deep Learning for Stiff Contact Dynamics}

\author{Mihir Parmar*, Mathew Halm*, and Michael Posa%
\thanks{*Co-first authors.}
\thanks{All authors are with the GRASP Laboratory, University of Pennsylvania, Philadelphia, PA 19104, USA. Correspondence to Mathew Halm
        {\tt\small mhalm@seas.upenn.edu}}%
}

\newcommand{\Real}{\mathbb{R}}

\begin{document}

\maketitle
\thispagestyle{empty}
\pagestyle{empty}

\begin{abstract}

Frictional contact has been extensively studied as the core underlying behavior of legged locomotion and manipulation, and its nearly-discontinuous nature makes planning and control difficult even when an accurate model of the robot is available.
Here, we present empirical evidence that learning an accurate model in the first place can be confounded by contact, as modern deep learning approaches are not designed to capture this non-smoothness.
We isolate the effects of contact's non-smoothness by varying the mechanical stiffness of a compliant contact simulator.
Even for a simple system, we find that stiffness alone dramatically degrades training processes, generalization, and data-efficiency.
Our results raise serious questions about simulated testing environments which do not accurately reflect the stiffness of rigid robotic hardware.
Significant additional investigation will be necessary to fully understand and mitigate these effects, and we suggest several avenues for future study.
\end{abstract}

\section{INTRODUCTION}
\label{sec:intro}
\begin{figure*}[!b]
	\vspace{-5mm}
	\begin{subfigure}{.10\hsize}
	\includegraphics[width=\hsize]{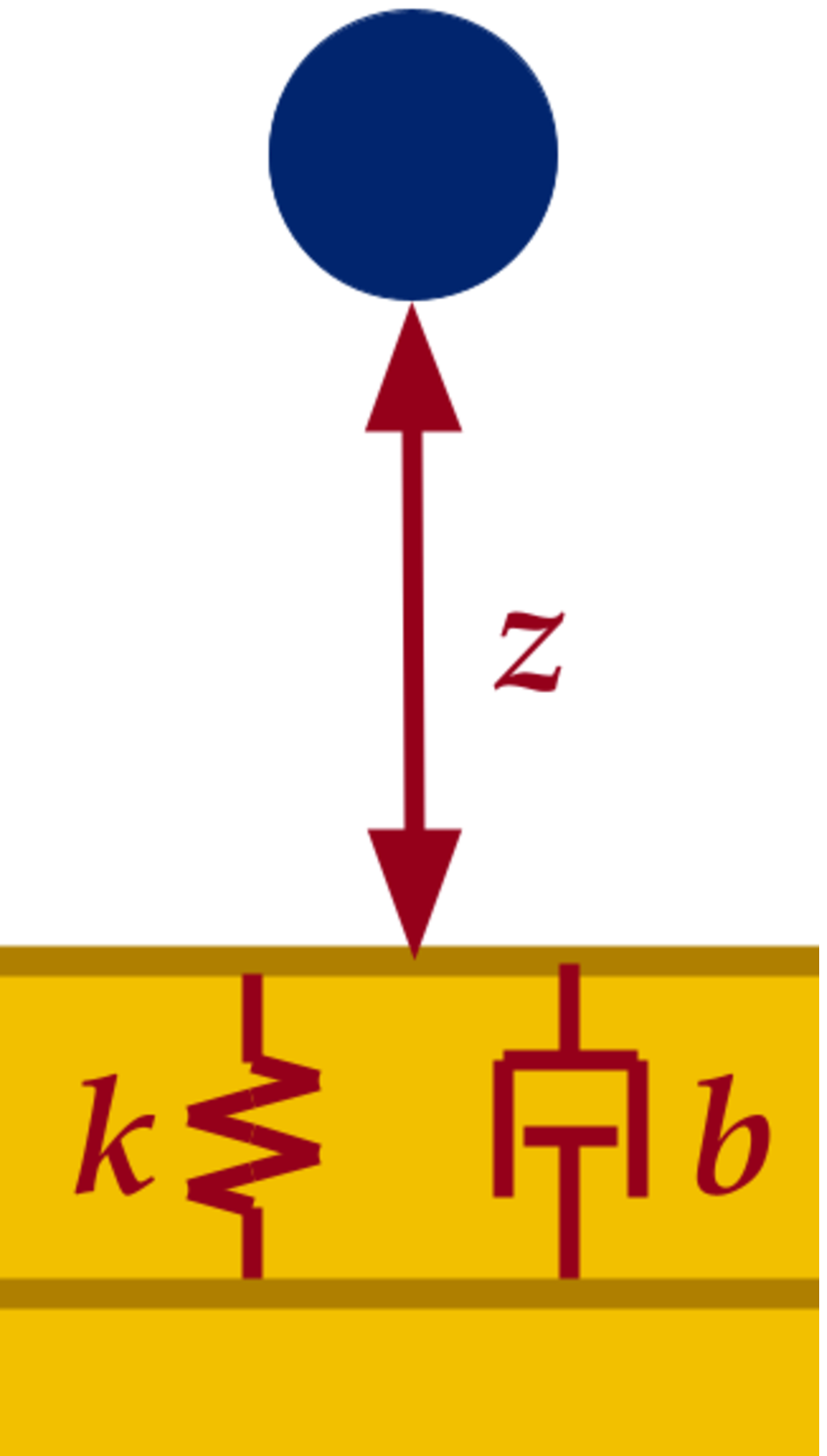}
	\caption{\label{subfig:1DDiagram}}
	\end{subfigure}
	\begin{subfigure}{.70\hsize}
	\centering
	\includegraphics[width=0.49\hsize]{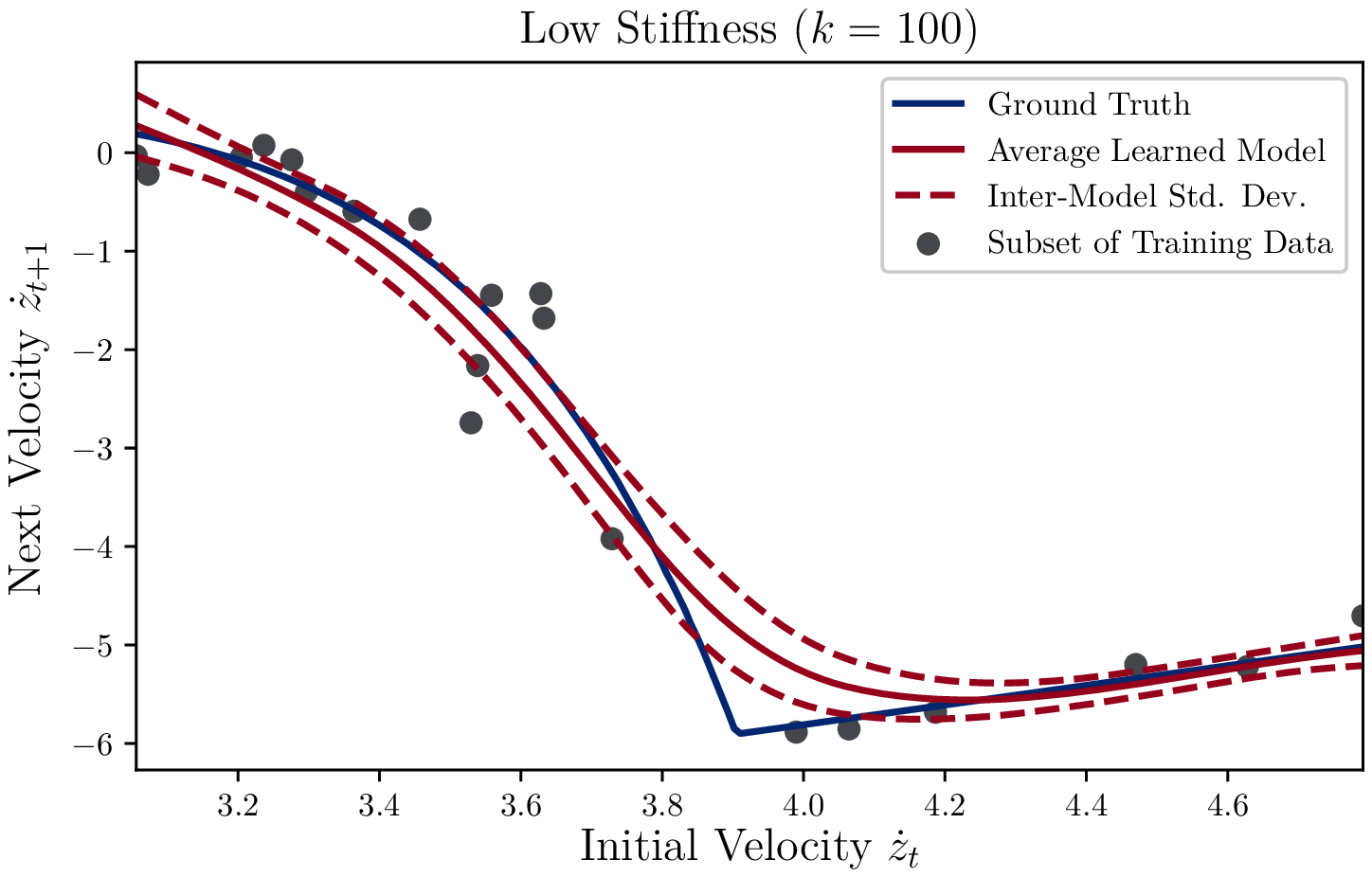}
	\includegraphics[width=0.49\hsize]{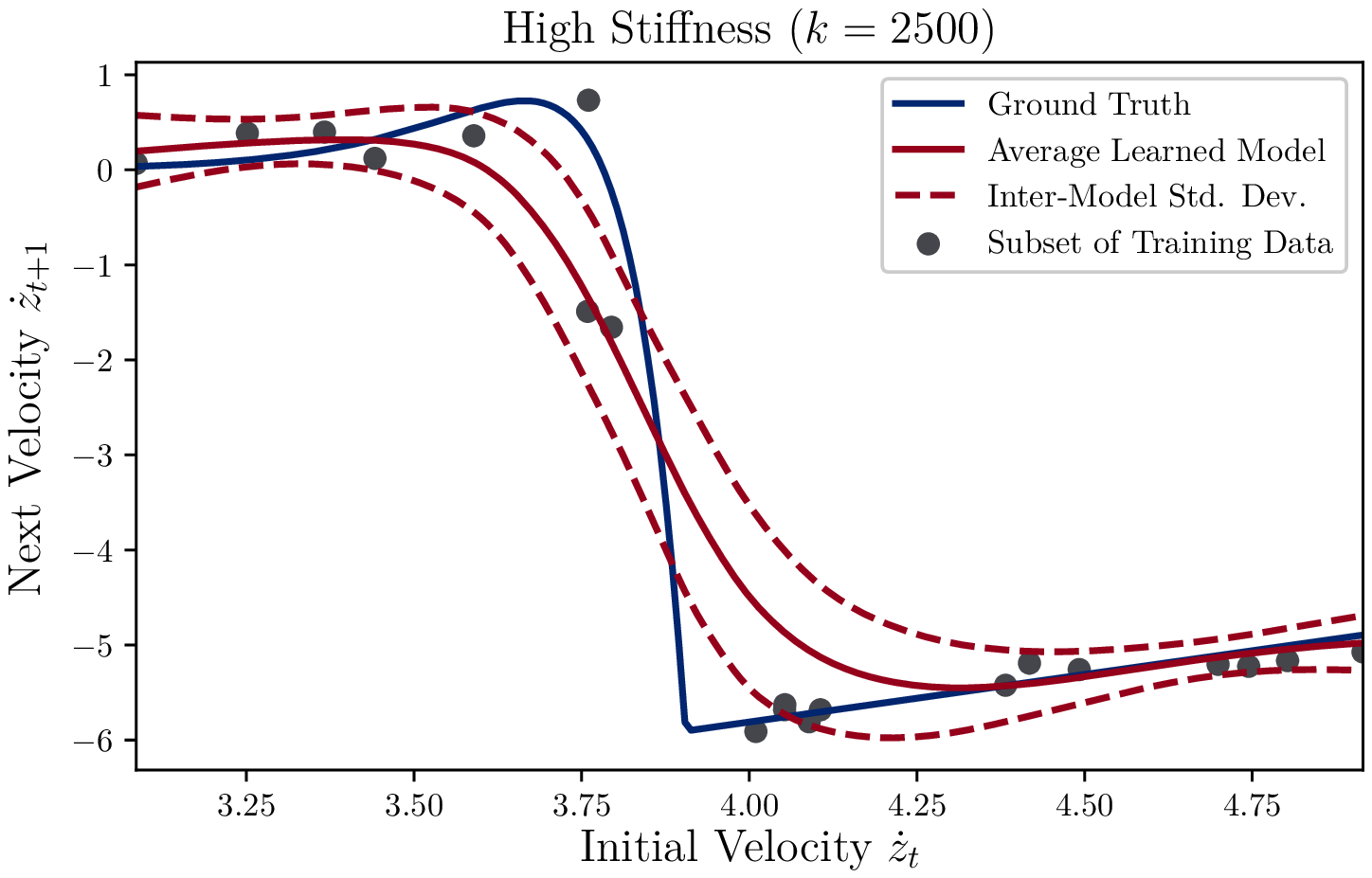}
	\vspace{-5mm}
	\caption{\label{subfig:1DModels}}
	\end{subfigure}
	\centering
	\caption{Challenges of learning stiff dynamics are shown on a 1-D example. (\subref{subfig:1DDiagram}): A point mass (blue) falls from an initial height $z_t = 1$ toward compliant ground (yellow), modeled as a spring-damper system. 
	(\subref{subfig:1DModels}): For each of two stiffnesses $(k)$, 100 predictive models are trained on noisy data to predict the next velocity $\dot z_{t+1}$ from different initial velocities $\dot z_t$.
	Learning performance is heavily degraded on the stiffer $k=2500$ system, despite meticulous hyperparameter optimization; training loss, ground-truth mean square error, and inter-model variance are 197\%, 413\%, and 309\% higher than for $k=100$.
	Details of this experiment may be found in Appendix \ref{adx:1DExample}.\label{fig:1DExample} }
\end{figure*}

Advances in model-based control and planning have always been essential to state-of-the-art robotic manipulation and locomotion.
Traditionally, roboticists have relied on high-accuracy, physics-based models of tightly-controlled laboratory environments.
However, as robotic systems transition to practical applications in unknown and unstructured environments, learning accurate models from limited data becomes increasingly important.
Despite recent successes leveraging modern deep learning to this end (e.g. \cite{Ajay2019,Janner2019,Chua2018,Nagabandi2019}), robotic performance remains decidedly sub-human in almost all scenarios.
One limiting factor is that contact is challenging to leverage, even given an accurate model; controlling the nearly-discontinuous behaviors of impact remains incredibly difficult, despite extensive study \cite{Wieber2016}.
In this work, we show the same properties also seriously impede common deep learning methods from obtaining an accurate model in the first place.

Impacts among robots and their surroundings are particularly difficult to model.
When objects collide, materials deform on an imperceptibly small spatial and temporal scale, preventing interpenetration.
The underlying material property driving this rapidity, \textit{mechanical stiffness}, causes multiple forms of \textit{numerical stiffness} in the equations of motion of these systems (See Figure \ref{fig:CubeIllustration}).
Slight inaccuracies in either initial conditions or model parameters can generate wildly-different predictions, even over a small time horizon \cite{Chatterjee1997,Ajay2018}.
Furthermore, measurements of the velocities are extremely sensitive to the time that they are recorded, as they change near-instantaneously during impact \cite{Fazeli2017Estimation,Halm2019}.
These properties become significant problems when learning a model of a real system from noisy sensor measurements.

Several associated issues have been duly noted in prior works on system identification (sysID) \cite{Kolev2015,Fazeli2017Estimation} and differentiable physics \cite{Heiden2020,LeLidec2021}.
In these settings, a handful of unknown parameters of a physics-based model are fit to data.
However, constructing an appropriate model space requires extensive knowledge of the robot and its surroundings; fitting model parameters requires expertise in both mechanics and optimization; and even excellent implementations are limited by inaccurate approximations, such as object rigidity and inelastic impact, inherent to tractable physical models \cite{Fazeli2017}.

The flexibility of deep neural networks (DNN's) can circumvent each of these issues, but they also introduce challenges \textit{unique} to their high-dimensional optimization setting.
First, stiffness in the learned dynamics introduce both stiffness and local minima into the training optimization landscape \cite{Kolev2015,Heiden2020}.
While sysID methods address this issue by exploiting either second-order \cite{Kolev2015} or global \cite{Heiden2020} optimization techniques, these tools cannot be tractably applied to the high-dimensional parameter spaces of DNN's.
Furthermore, while a fundamental advantage of DNN's is their ability to approximate \textit{any} dynamical system, a corresponding challenge is that \textit{many} distinct model parameters may fit the training data well, and one must be selected via an inductive bias.
Unfortunately, in direct conflict with the stiff and nearly-discontinuous behaviors of frictional contact, deep learning techniques tend to select the smoothest interpolator of the data.
This behavior is both a predisposition of common training techniques like stochastic gradient descent (SGD) \cite{Belkin2019,Ribeiro2020} and an explicit goal of common regularizers such as weight-decay and spectral normalization \cite{Zhang2018,Miyato2018}.
This smoothing effect is particularly harmful when available data is sparse, as can be see in a 1D example in Figure \ref{fig:1DExample}.
If stiffness indeed significantly influences the performance of deep learning, then serious questions must be raised not only about robotic learning methods, but also the relevance of simulated results to the robotics community.
Many simulators (e.g. MuJoCo \cite{Todorov2014}) allow users to specify mechanical stiffness; properly used, simulation can produce physically-accurate behaviors \cite{Kolev2015,Erez2015}.
However, ubiquitous benchmarking suites often use software default values for stiffness, rather than values tuned for realism \cite{Brockman2016}.
The idea that unrealistic contact settings can generate a gap between simulated and real-world performance is an existing intuition in the robotics community \cite{Tedrake}.
However, this phenomenon has not been rigorously examined in the literature.

In this paper, we contribute an empirical quantification and isolation of the detrimental effects of stiffness on deep learning performance.
We begin by describing how stiffness enters into the equations of motion of a simple simulated system in MuJoCo in Section \ref{sec:system}, and show that default settings are significantly less stiff than many real systems.
In Section \ref{sec:experiments}, we propose a testing methodology to examine the negative effects of stiffness on inherent unpredictability, training process degradation, generalization, and long-term prediction.
As is common intuition, our results (Section \ref{sec:results}) show that raising stiffness degrades ground-truth model predictions as the underlying system becomes more sensitive to noise.
However, we find that stiffness induces multiple pathological behaviors \textit{beyond} this effect:
\begin{enumerate}
	\item The training error of learned models degrades with stiffness nearly \textit{twice} as fast as the ground-truth model, even for single-step predictions.
	\item While generalization error can be eliminated for non-stiff systems with ample training data, test error stays significantly higher than training error for stiff systems.
	\item Data-efficiency degradeds $100$-fold for our stiffest models when evaluated on long-term prediction. 
\end{enumerate}
These results raise two serious questions which we encourage the robotics community to confront head-on: \textit{Are we correctly utilizing deep learning's most powerful and essential behaviors in our current methods? And do our simulated environments faithfully capture essential challenges of real-world phsyics?}
In Section \ref{sec:conclusion}, we list current research related to these questions (including our prior work \cite{Pfrommer2020}), and furthermore list several unaddressed challenges.

\section{EXAMPLE SYSTEM}
\label{sec:system}
We now describe a simple example system and associated data generation methodology, for which we will isolate the effects of stiffness on learning performance.

\subsection{Simulation Environment}
While the many uncontrollable factors of real-world experiments offer a challenging environment to test newly-developed algorithms' performance, the primary goal of this paper is to \textit{isolate} the effects of stiffness on commonplace methods in robotics.
Unmodeled material complexities and unknowable noise distributions in a real robotic system would therefore befuddle the results presented here, rather than strengthen them.
We therefore conduct our experiments in a simulated environment, which allows them to be easily repeated or used as a benchmarking task in future research.
We conduct our experiments MuJoCo \cite{Todorov2014}, because it allows for direct control over contact stiffness.
Using MuJoCo also enhances the relevance of our results to ubiquitous benchmarking suites which use the simulator \cite{Brockman2016,Tassa2020}.

While MuJoCo models objects as being exactly rigid, it allows for an interpretation of mechanical stiffness by using a ``soft contact'' model similar to the one used in Figure \ref{fig:1DExample}; a detailed discussion can be found in \cite{Todorov2014} and in the online MuJoCo documentation\footnote{\url{http://www.mujoco.org/book/computation.html}}. 
MuJoCo solves for appropriate contact forces with a convex optimization problem, and thus there is no closed form expression for the forces as a function of the current state.
However, when an object makes contact with a static environment, inter-body penetration $r$ approximately\footnote{A more detailed treatment of this behavior can be found in Appendix \ref{adx:DataCollectionDetails}.} obeys
\begin{equation}
	\ddot r \approx -b\dot r - kr\,.
	\label{eq:ApproximateContactDynamics}
\end{equation}
Here, the ``stiffness'' $k$ is the primary mechanism resisting penetration, and the damping ratio $\zeta = \frac{b}{2\sqrt{k}}$ controls elasticity of impacts.
Similar techniques have long been used for stable simulation of constrained dynamical systems, dating back to Baumgarte's 1972 formulation \cite{Baumgarte1972}.
We note that the units of $k$ are \si{\newton\over\kilogram\meter}, whereas mechanical stiffness is typically expressed in \si{\newton\over\meter} units. 
MuJoCo's default values for $k$ are in the $2000$--$2500$\si{\newton\over\kilogram\meter} range.
However, as we will discuss in Section \ref{subsec:DataGeneration}, the corresponding contact behavior is far softer than that of many real-world objects, including common robotic platforms.

\begin{figure}
	\includegraphics[width=0.8\hsize]{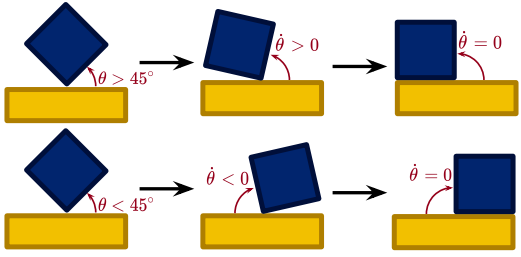}
	\centering
	\caption{Sensitivity to initial conditions and near-instantaneous impact of a 2D block on flat ground are shown. (left) Two trajectories begin from nearly identical initial conditions, where the block (blue) contacts the ground (yellow) at 1 corner; the center of mass is left of the contact point in the upper trajectory and to the right in the lower one. (center) after some time has elapsed, the state of the cube differs drastically in the two trajectories. (right) the velocity of the cube jumps to zero abruptly as it hits the ground in both cases.\label{fig:CubeIllustration}}
\end{figure}
\subsection{System Description}
High-dimensional systems, much like real-world environments, are also commonly used to stress-test new robot learning algorithms \cite{Brockman2016}. 
By contrast, in this paper, picking a simple, low-dimensional system instead allows us to more thoroughly and tractably analyze the effects of stiffness under reduced computational and sample complexity.
We follow previous studies (\cite{Jiang2018,Fazeli2020,Pfrommer2020}), and choose a ``die roll'' system, in which a single, rigid cube makes contact with the ground.
Despite being low-dimensional, the cube exhibits many of the hallmark challenges in contact modeling: stick-slip transition, discontinuous impact, multiple contact points, and extreme sensitivity to initial conditions; Figure \ref{fig:CubeIllustration} illustrates some of these behaviors in 2D.

Our 3D die system has a 13-dimensional state
\begin{equation}
	x_t = \begin{bmatrix}
		p_t\;; & q_t \;; & \dot p_t \;; & \omega_t
	\end{bmatrix}\,,
\end{equation}
where $p_t \in \Real^3$ is the center of mass position; $q_t \in S^3$ is the orientation of the cube, expressed as a quaternion; $\dot p_t \in \Real^3$ is the world-frame c.o.m. velocity; and $\omega_t \in \Real^3$ is the body-frame angular velocity.
We will often identify the generalized velocity of the system as $v_{t} = [\dot{p}_{t};\; \omega_{t}]$.
When simulating the dynamics in discrete-time, Newton's second law is often approximated with a semi-implicit formulation, as in MuJoCo \cite{Todorov2014}.
These equations have the form
\begin{equation}
	x_{t+1} = f(x_t)\,.\label{eq:DynamicalSystem}
\end{equation}
For a symmetric cube\footnote{the symmetry assumption implies that the Coriolis forces are zero, as the inertia tensor is a multiple of the identity matrix.}, $f$ is calculated as
\begin{align}
	m(\dot p_{t+1} - \dot p_t) &= (F(x_t)-mg)\Delta t\,,\label{eq:NewtonEquation} \\
	I(\omega_{t+1} - \omega_{t}) &= \tau(x_t)\Delta t\,,\label{eq:EulerEquation} \\
	{p_{t+1} -  p_t} &= \dot p_{t+1}\Delta t\,,\label{eq:PositionIntegration} \\
	q_{t}^{-1} \otimes q_{t+1} &= Q(\omega_{t+1}\Delta t)\label{eq:OrientationIntegration} \,,
\end{align}
where $\Delta t$ is the time-step duration; $m$ and $I$ are the cube's mass and inertia; $g$ is the gravitational acceleration vector; and $F$ and $\tau$ are the average contact force and torque over the time step.
$Q(v) = [\cos\frac{||v||_2}{2}; \hat v \sin\frac{||v||_2}{2}]$ is the quaternion corresponding to a rotation of angle $||v||_2$ about axis $\hat v = \frac{v}{||v||_2}$ and $\otimes$ is the quaternion product.
We use system parameters that are identical to the real system used in \cite{Pfrommer2020}; a full list can be found in Table \ref{table:SystemParameters}.
\begin{table}
 \caption{Die Roll System Parameters}
\label{table:SystemParameters}
\centering
\begin{tabularx}{0.7\linewidth}{@{}l*{10}{c}c@{}}
\toprule
Constant & Symbol & Value & Units \\ 
\midrule
mass & $m$ & $0.37$ & \si{\kilogram} \\ 
inertia & $I$ & $\num{6.167e-4}$ & \si{\kilogram\meter\squared} \\ 
side length & $l$ & $0.1$ & \si{\meter} \\
gravity & $g$ & $9.81$ & \si{\meter\over\second\squared}\\
friction coefficient & $\mu$ & $1$ & (none) \\
stiffness & $k$ & (varies) & \si{\newton\over\kilogram\meter} \\
damping ratio & $\zeta$ & $1.04$ & (none) \\
time-step & $\Delta t$ & $\num{6.74e-3}$ & \si{\second}\\

\bottomrule
\end{tabularx}
\vspace{-1mm}
\end{table}
\subsection{Data Generation}\label{subsec:DataGeneration}
In order to isolate how different stiffnesses $k$ generate different behaviors, for each of 3 stiffnesses listed in Table \ref{table:StiffnessSettings}, we generate a dataset $\{\tau\}$ of ``dice roll'' trajectories $\tau = \{x_{0}, x_{1}, x_{2},...x_{T-1}\}$  with an identical process, summarized here and detailed further in Appendix \ref{adx:DataCollectionDetails}. 
We refer to the three stiffnesses as \textit{Hard}, \textit{Medium} and \textit{Soft}.

We instantiate the system for a given stiffness in MuJoCo with the parameters in Table \ref{table:SystemParameters}; the damping coefficient $b$ is selected to keep the damping ratio $\zeta$ consistent between stiffnesses.
Initial states are sampled uniformly around a nominal state $x_{0,ref}$.
From the initial state, we simulate forward in time with MuJoCo's dynamics \eqref{eq:NewtonEquation}--\eqref{eq:OrientationIntegration} until the cube impacts the ground and comes to rest.

In the real world, position and orientation of similar systems are commonly tracked via computer vision, which can incur a small amount of slowly-drifting measurement noise \cite{Pfrommer2020}.
To approximate this error, for each trajectory, we add a small, uniformly random offset to the entire trajectory, and a second round of smaller noise independently to each datapoint.
Finally, velocity states are reconstructed using the finite difference equations \eqref{eq:PositionIntegration}--\eqref{eq:OrientationIntegration} on the noisy configurations.
The total noise injected through this process on each state variable is on the order of $1$ \si{\milli\meter}, \si{\deg}, \si{\milli\meter\over\second}, or \si{\deg\over\second}.

\begin{figure}[ht]
	\centering
	{\includegraphics[width=0.32\hsize]{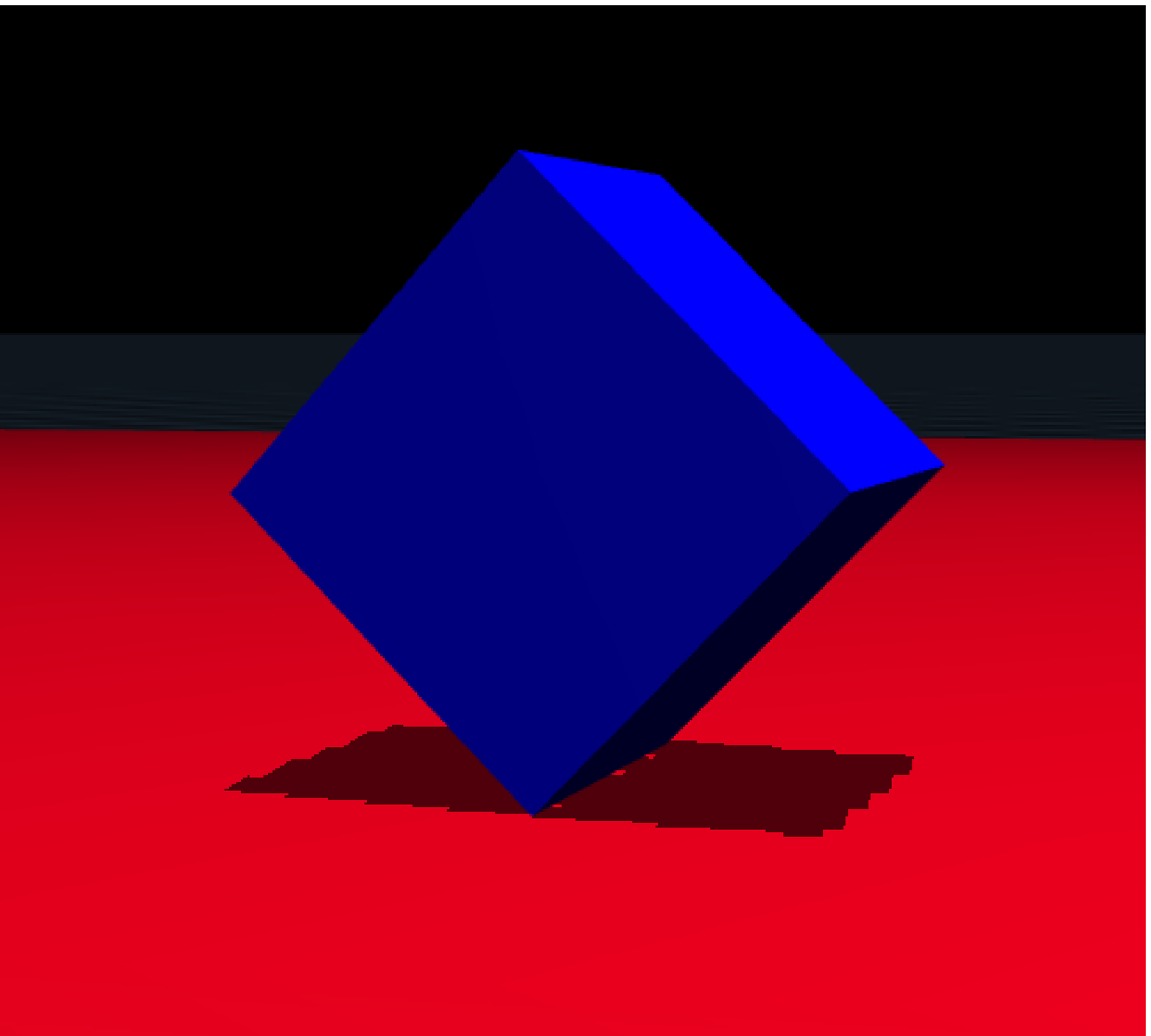}}
	{\includegraphics[width=0.32\hsize]{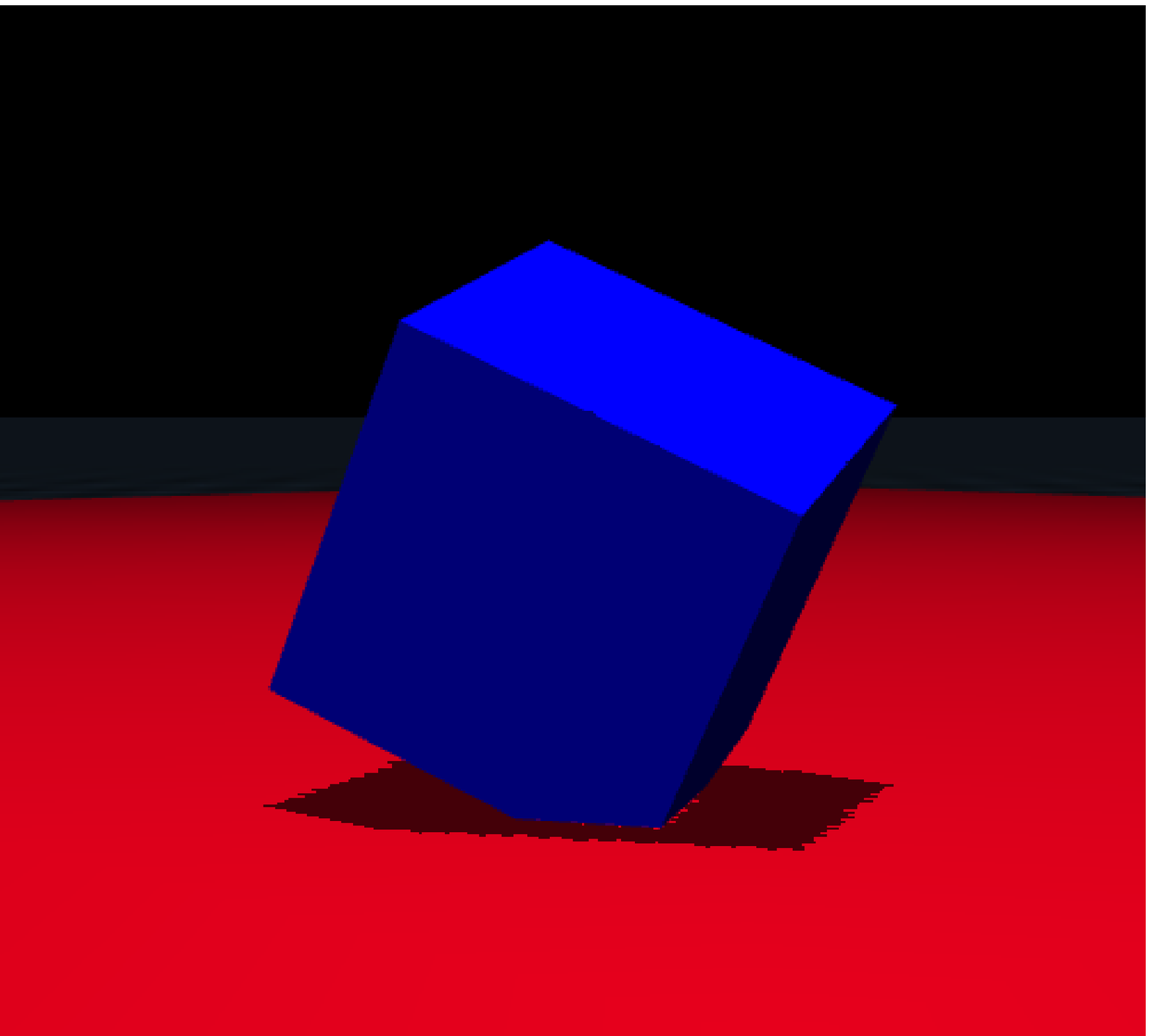}}
	{\includegraphics[width=0.32\hsize]{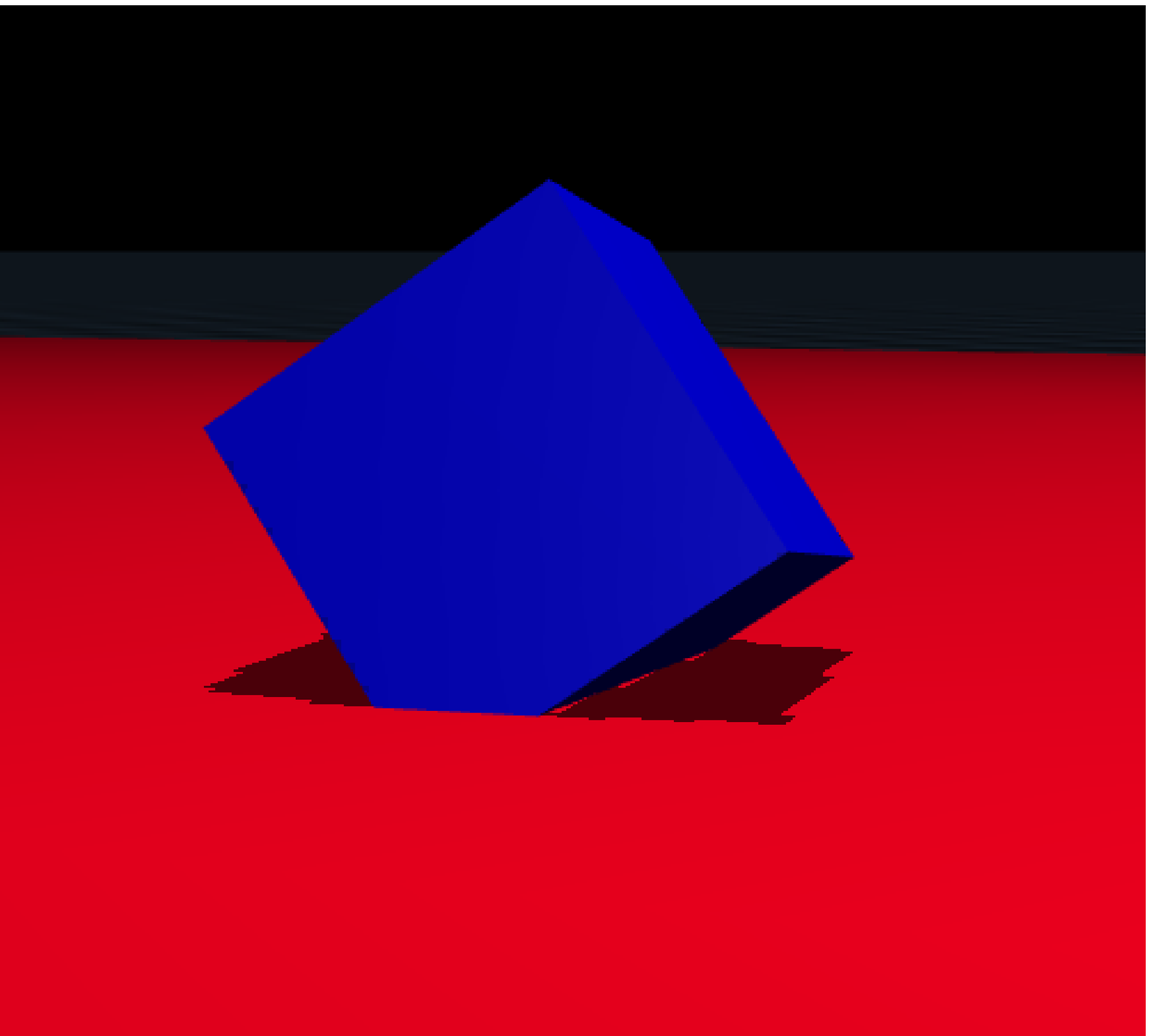}}

	\caption{From left to right, illustration from MuJoCo demonstrating the variation in the amount of ground penetration for \textit{Hard}, \textit{Medium}, and \textit{Soft} settings, respectively. These visualizations are captured from trajectories with identical initial states. \label{fig:PenetrationSnaps} }
	\vspace{-3mm}
\end{figure}

For each stiffness setting, we collect $10,000$ trajectories $\{\tau\}_{\text{train}}$ for hyperparameter optimization and training purposes, and $1,000$ more trajectories $\{\tau\}_{\text{eval}}$ for evaluation of the optimized models.
To evaluate physical realism of each of these settings, we also compute the maximum ground penetration of the die, averaged over trajectories (Table \ref{table:StiffnessSettings} and Figure \ref{fig:PenetrationSnaps}).
Even for the \textit{Hard} stiffness, which is comparable to MuJoCo's default, we observe ground penetration of around $10\%$ of the die body-length.
By comparison, deformations on real-world objects are often imperceptible to the human eye.
Thus, even our \textit{Hard} model is far less stiff than the real-world dynamics upon which our system was based \cite{Pfrommer2020}.

\begin{table}
	\caption{Stiffnesses and corresponding ground penetrations}
	\label{table:StiffnessSettings}
	\centering
	\begin{tabularx}{0.77\linewidth}{c | c | c}
		\toprule
		Stiffness Setting & $k$ $\big(\si{\newton\over\kilogram\meter}\big)$ & Max. Penetration (\si{\milli\meter}) \\ 
		\midrule
		\textit{Hard} & 2500  & 12\\ 
		\textit{Medium} & 300  &  26 \\ 
		\textit{Soft} & 100  &  40 \\
		
		\bottomrule
	\end{tabularx}
	\vspace{0mm}
\end{table}

\section{EXPERIMENTS}
\label{sec:experiments}
We now describe the process of learning a dynamical system from data; challenges that stiffness imposes in dynamics learning; and motivate the design of our experiments.
The associated Pytorch codebase is available online\footnote{\url{https://github.com/DAIRLab/ContactLearningBias}}.
\subsection{Representing System Dynamics with Neural Networks}
Many modern deep learning-based modeling approaches (e.g. \cite{Ajay2019,Janner2019,Chua2018,Nagabandi2019,Ajay2018}) follow the same fundamental approach: fitting a neural network approximation of the system dynamics \eqref{eq:DynamicalSystem} directly to data.
In these methods, a DNN ${f}_\theta\left({x_{t-h+1:t}}\right)$ with parameters $\theta$ outputs the next-state $x_{t+1}$ given a window of previous states $x_{t-h+1:t}$, where $h$ is the history length.
In this paper, we implement two of the most common architectures.
The first, and most elementary, is to pick $h=1$ and map $x_t$ to $x_{t+1}$ with a simple multilayer perceptron (MLP) (\cite{Janner2019,Nagabandi2019}).
However, this method has been shown to struggle with noisy data in a manipulation setting \cite{Ajay2019,Ajay2018}, an expected behavior given the sensitivity of contact dynamics \cite{Kolev2015}.
An intuition in robotics is that better estimates of the current state can be generated by fusing multiple sensor readings.
Correspondingly, a common approach is to use recurrent neural networks (RNN's) with history length $h>1$ \cite{Ajay2019,Ajay2018}, and thus our second set of architectures are RNN-based.
We experiment with the three RNN variants: Long Short Term Memory (\textit{LSTM}) \cite{LSTM}, Gated Recurrent Unit (\textit{GRU}) \cite{cho2014learning} and Bi-directional LSTM (\textit{BiLSTM}) \cite{BiLSTM}.

Internally, MuJoCo predicts the next velocity $v_{t+1}$, and then constructs the next configuration using the finite-differencing method given by \eqref{eq:PositionIntegration}-\eqref{eq:OrientationIntegration} \cite{Todorov2014}. 
Since our velocity data is generated with the same finite-difference, it is sufficient to output either ${f}_\theta\left({x_{t-h+1:t}}\right) \approx v_{t+1}$ or ${f}_\theta\left({x_{t-h+1:t}}\right) \approx \Delta{v} = v_{t+1} - v_{t}$ from the network, and then reconstruct the next configuration with \eqref{eq:PositionIntegration}-\eqref{eq:OrientationIntegration} as MuJoCo does.
The empirically determined optimal network structure and target variable choice for all stiffness setting were found to be GRU and $v_{t+1}$ respectively, as listed in Table \ref{table:OptimalHyperparameters}.
A detailed explanation of the process is given in Appendix \ref{adx:LearningDetails}.

\subsection{Training Process}
To train one of our networks, we first aggregate a set of $N$ trajectories $\{\tau_{1:N}\}$ randomly sampled from $\{\tau\}_{\text{train}}$ and slice them into training data inputs $ \{x_{t-h+1:t}\}$ and corresponding outputs $\{v_{t+1}\}$.
To improve numerical conditioning during training, we follow a standard procedure of normalizing the input data to have zero mean and unit variance \cite{Nagabandi2019}.
We further split the sliced data $ \{x_{t-h+1:t},v_{t+1}\}$ in 70:20:10 proportions into training (${D}_{\text{train}}$), validation (${D_{\text{val}}}$) and test (${D_{\text{test}}}$) sets.

For a dataset $D$ with $|D|$ observations, we define the mean-square error loss over $D$ for a model $f$ as
\begin{equation}
	\mathcal{L}(f, D)=\frac{1}{|D|} \sum_{\left({x_{t-h+1:t}}, {v_{t+1}}\right) \in {D}} \left\|\left({v}_{t+1}\right)-{f}\left({x_{t-h+1:t}}\right)\right\|_{2}^{2}\,.
	\label{eq:TrainingError}
\end{equation}
Accordingly, we train our dynamics models using the Adam optimizer \cite{kingma2017adam} to minimize $\mathcal L(f_\theta, D_{\text{train}})$.
We terminate training with early stopping with a patience of $30$ epochs, and save the model with the lowest validation loss $\mathcal L(f_\theta, D_{\text{val}})$. Test set error $\mathcal L(f_\theta, D_{\text{test}})$ is then used as the metric during hyperparameter optimization.
To provide an optimistic perspective on how the dynamics of each stiffness setting can be learned, we optimize separate hyperparameters for each stiffness.
This process is detailed in Appendix \ref{adx:LearningDetails}, and
Table \ref{table:OptimalHyperparameters} specifies the final set of selected hyperparameter values for each of the stiffness setting.

\begin{table}
	\caption{Optimized Hyperparameters}
	\label{table:OptimalHyperparameters}
	\centering
	\begin{tabularx}{0.70\linewidth}{c c c c}
		\toprule
		\multirow{2}{*}{Hyperparameter} & \multicolumn{3}{c}{Stiffness Setting} \\
		& \textit{Hard} & \textit{Medium} & \textit{Soft} \\
		\midrule
		Network architecture & GRU & GRU & GRU \\
		Target variable & $v_{t+1}$ & $v_{t+1}$ & $v_{t+1}$\\
		learning-rate & 1e-4 & 1e-5 & 1e-5 \\ 
		hidden-size & 128 & 128 & 128 \\
		history-length & 16 & 16 & 16 \\
		weight-decay & 0 & 4e-5 & 4e-5 \\
		\bottomrule
	\end{tabularx}
	\vspace{-7mm}
\end{table}

\subsection{Measuring Stiffness's Effect on Learning Performance}

To perform an optimistic analysis on how well learning algorithms perform on systems with different stiffnesses, we focus performance analysis on our hyperparameter-optimized models in three settings: the effectiveness of Adam in minimizing the training set loss; generalization performance; and long-term prediction performance.

Broadly, the goal of supervised learning methods, including our dynamics learning process, is to generate accurate outputs for unseen inputs.
In deep learning, this behavior is quantified as low test error $\mathcal L(f_\theta, D_{\text{test}})$.
Our discussion in Section \ref{sec:intro} suggests that the test error will increase for higher stiffness settings, but it is important to note that the test set error can be driven up by many mechanisms.
Inspired by the separate treatment of approximation, estimation, and generalization error common in statistical learning theory \cite{Poggio2003}, we now define, motivate, and hypothesize about the following decomposition of the test error:
\begin{align}
	\mathcal L(f_\theta, D_{\text{test}}) &= \mathcal L(f_{oracle}, D_{\text{train}})\nonumber \\
	 &\quad+ (\mathcal L(f_{\theta}, D_{\text{train}}) - \mathcal L(f_{oracle}, D_{\text{train}}))\label{eq:testdecomposition} \\
	 &\quad+ (\mathcal L(f_\theta, D_{\text{test}}) - \mathcal L(f_\theta, D_{\text{train}}))\,.\nonumber
\end{align}

As discussed in Section \ref{sec:intro}, it is important to acknowledge that even if the \textit{exact} MuJoCo model used to generate the data is simulated from a noisy initial condition, it will make an imperfect prediction of the future states.
Furthermore, it is well known that stiff dynamics will exacerbate this issue \cite{Chatterjee1997,Kolev2015,Pfrommer2020}.
We capture this effect with the first term in the test error decomposition \eqref{eq:testdecomposition}, $e_{oracle} = \mathcal L(f_{oracle}, D_{\text{train}})$.
To be precise, our \textit{MuJoCo oracle} model $f_{oracle}(x_{t-h+1:t})$ predicts the next velocity $v_{t+1}$ with MuJoCo using the current state $x_t$ and the Netwton-Euler equations \eqref{eq:NewtonEquation}-\eqref{eq:EulerEquation}.
As the \textit{MuJoCo oracle} captures the underlying true behavior of the system, it serves as a natural, optimal baseline for the learned models.

Since prediction loss can be poorly conditioned when learning stiff dynamics \cite{Heiden2020,Pfrommer2020}, we hypothesize that Adam will have difficulty in converging to good minima consistently; therefore, the training loss at convergence of stiffer models is likely to have higher mean.
To test this hypothesis, we train models on data with different stiffnesses and dataset sizes, and then observe how much the resulting training error is degraded in comparison to the performance of the MuJoCo oracle: $(\mathcal L(f_{\theta}, D_{\text{train}}) - \mathcal L(f_{\text{oracle}}, D_{\text{train}}))$.

Deep learning is biased towards fitting a smooth interpolator on the data \cite{Belkin2019,Ribeiro2020}; however, as we note in Section \ref{sec:intro}, the underlying behavior of contact is non-smooth.
Hence, at equal training set sizes, we expect that DNNs fit to stiffer systems' data will suffer worse generalization error.
We examine this hypothesis by comparing the generalization error $(\mathcal L(f_\theta, D_{\text{test}}) - \mathcal L(f_\theta, D_{\text{train}}))$ of learned models corresponding to different stiffnesses and dataset sizes.

While we have followed a commonly used approach by training our models on single-step predictions (e.g. \cite{Chua2018,Nagabandi2019}), long-term prediction quality is essential for model-based control methods, such as MPC \cite{Chua2018}.
We therefore additionally evaluate our models' long-term prediction capability.
For a particular ground-truth trajectory $\tau \in \{\tau\}_{\text{eval}}$, we use the initial $(h)$ ground-truth states  $\{x_{t}\}_{t=0}^{t = h-1} \in \tau$ as input for the learned model and recursively construct predicted trajectory for next $\hat{T}$ time-steps $\{\hat{x}_{t}\}_{t=h}^{t = h+\hat{T}-1}$.
Similar to \cite{Pfrommer2020}, we report the temporally-averaged absolute position and rotation error for each model:
\begin{align*}
	e_{pos}&=\frac{1}{\hat{T}} \sum_{j=h}^{h+\hat{T}-1}\left\|{\boldsymbol{\hat{p}}}_{j}-\boldsymbol{p}_{j}\right\|_{2}\,, &
	e_{rot}&=\frac{1}{\hat{T}} \sum_{j=h}^{h+\hat{T}-1}\left|\operatorname{angle}\left(\boldsymbol{\hat{q}}_{j}, {\boldsymbol{q}}_{j}\right)\right|\,,
\end{align*}
where, $\operatorname{angle}(\cdot,\cdot)$ represents the relative angle between two quaternions.

To make a fair evaluation for models with different history length and simultaneously ensure that the prediction horizon is long enough to capture ground impact and block tumbling, we use $\hat{T} = 50$ (a $337$ \si{\milli\second} duration) in our experiments.
\section{RESULTS}
\label{sec:results}
In Table \ref{table:OracleSinglestepErrors} we report \textit{MuJoCo oracle}'s performance on the single-step velocity prediction $(\mathcal L(f_{oracle}, D_{\text{train}}))$ task across different stiffnesses.
As expected, we observe that the single-step prediction performance of the \textit{MuJoCo oracle} improves as the contact is made softer.

\begin{table}[!h]
	\caption{\textit{MuJoCo oracle} Prediction Performance}
	\label{table:OracleSinglestepErrors}
	\centering
	\begin{tabularx}{\linewidth}{c c c c}
		\toprule
		Stiffness Setting & $\mathcal L(f_{oracle}, D_{train})$ & $e_{pos}$ (\% width) & $e_{rot}$ (\si{\deg}) \\
		\midrule
		\textit{Hard} & 0.0836 $\pm$ \num{1.8e-3} & $4.31 \pm 0.13$ & $3.98 \pm 0.03$\\
		\textit{Medium} & 0.011 $\pm$ \num{1.8e-4} & $3.57 \pm 0.08$ & $3.26 \pm 0.03$\\
		\textit{Soft} & \num{3.19e-3} $\pm$ \num{1.9e-5} & $2.92 \pm 0.04$ & $2.77 \pm 0.03$\\ 
		\bottomrule
	\end{tabularx}
	\vspace{0mm}
\end{table}

\begin{figure*}[t]
	\centering
	\begin{subfigure}{.35\hsize}
	   \includegraphics[width=\hsize]{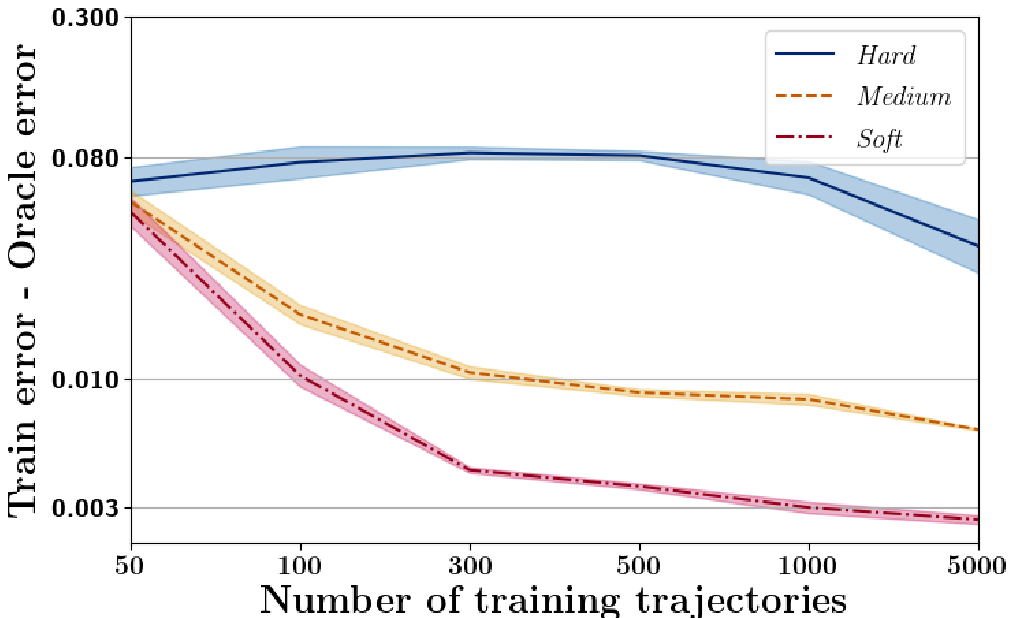}
       \caption{\label{subfig:train}}
    \end{subfigure}
    \hspace{-6.5mm}
    \begin{subfigure}{.35\hsize}
	   \includegraphics[width=\hsize]{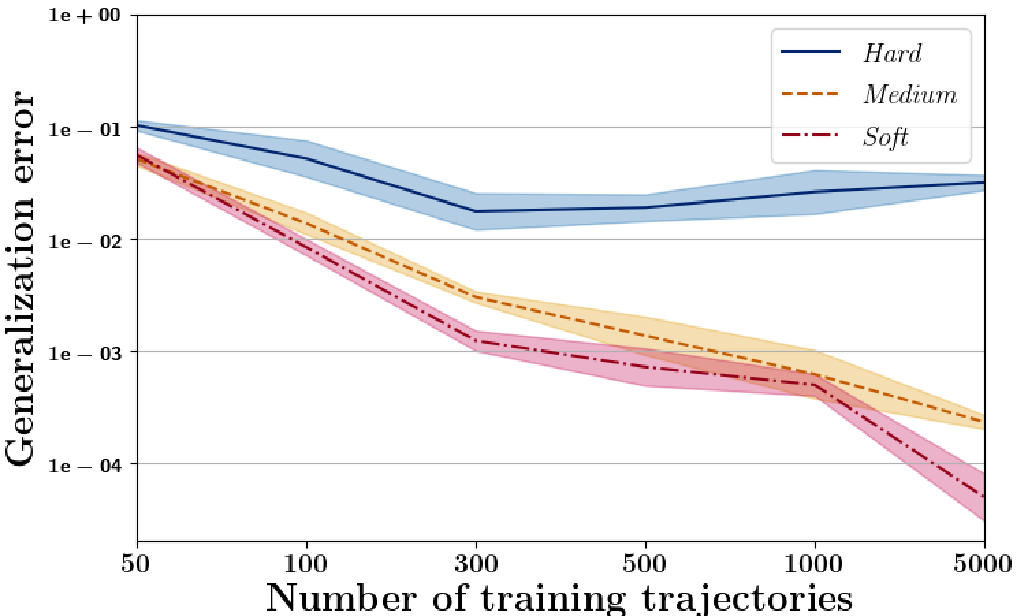}
       \caption{\label{subfig:generalization}}
    \end{subfigure}
    \hspace{-7mm}
    \begin{subfigure}{.35\hsize}
	   \includegraphics[width=\hsize]{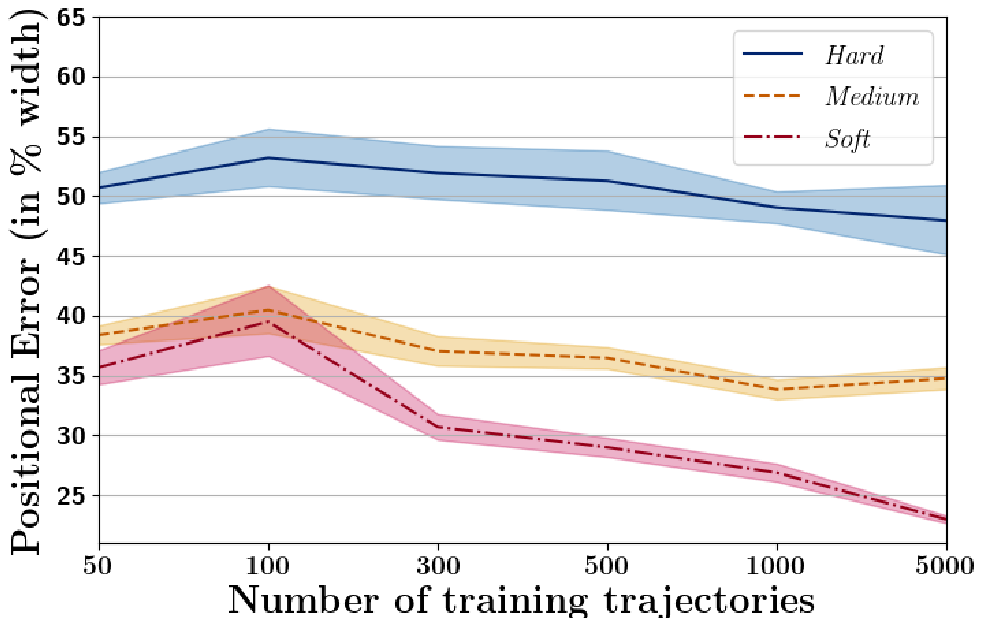}
       \caption{\label{subfig:positionerror}}
    \end{subfigure}
    \begin{minipage}[c]{0.63\hsize}
    \caption{We compare the quality of learned models across stiffness settings and dataset sizes. We plot several metrics with $95$\% log-normal confidence intervals. (\subref{subfig:train}): As stiffness increases, the gap between the training error and the oracle's error grows. (\subref{subfig:generalization}): Stiffer models show a significantly higher average generalization error across dataset sizes. (\subref{subfig:positionerror},\subref{subfig:rotationerror}): We compare the performance of our optimized networks on long-term prediction of position and orientation. Data-efficiency of learning \textit{Hard} models is at least $100$x worse than for \textit{Soft} models.
	}
  \end{minipage}
  \begin{minipage}[c]{0.35\hsize}
    \begin{subfigure}{\hsize}
	   \includegraphics[width=\hsize]{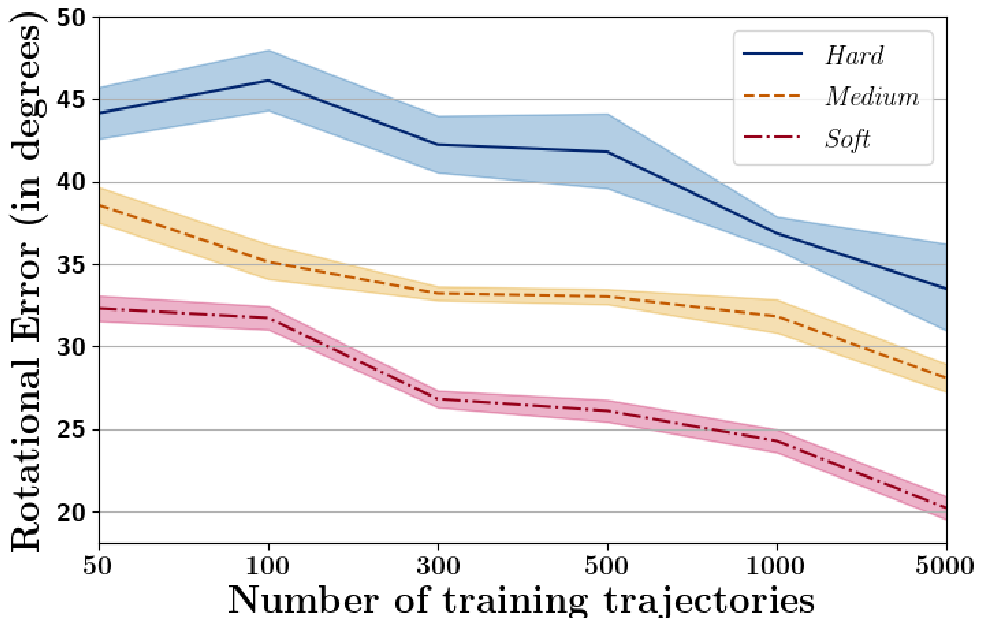}
       \caption{\label{subfig:rotationerror}}
    \end{subfigure}
	\end{minipage}
	\vspace{0mm}
\end{figure*}

While the single-step oracle error for \textit{Hard} contact is nearly $20$x higher when compared to \textit{Soft} contact, we find that it only accounts for approximately half of the training error.
Fig. \ref{subfig:train} demonstrates the difference in the converged training loss of learned models and the oracle single-step errors $(\mathcal L(f_{\theta}, D_{\text{train}}) - \mathcal L(f_{oracle}, D_{\text{train}}))$ across different dataset sizes and different contact settings.
The resulting values were right skewed and non-negative; therefore, we assume their distribution to be log-normal and construct its $95\%$ confidence interval using Cox's method \cite{Land1972}.
Stiffer models show worse average training loss across all tested dataset sizes with high confidence, with nearly $10$x gap between the \textit{Hard} and \textit{Soft} models in the high data regime.
Furthermore, there is a large data efficiency gap; \textit{Hard} models trained on $5000$ trajectories perform worse than both \textit{Medium} and \textit{Soft} models trained at just $100$ trajectories.

In Fig. \ref{subfig:generalization}, we plot the generalization error of the learned models with their $95\%$ log-normal confidence intervals. 
Similar to the trends noted in Fig. \ref{subfig:train}, we observe that the generalization error also exhibits a $10$x gap in the high data regime, and that again \textit{Hard} models perform worse than their \textit{Medium} and \textit{Soft} counterparts do with $50$x less data.
Additionally, while \textit{Medium} and \textit{Soft} models improve generalization by over a factor of $100$ as the dataset size increased from $50$ to $5000$ trajectories, \textit{Hard} models by contrast improve by less than a factor of $10$.

We capture the long-term prediction errors ($e_{pos}$, $e_{rot}$) of the \textit{MuJoCo oracle} in Table \ref{table:OracleSinglestepErrors} and that of the learned models in Fig. \ref{subfig:positionerror}--\ref{subfig:rotationerror}.
The MuJoCo oracle errors are at least $10$x smaller than the learned models for both metrics.
We also observe that for both \textit{MuJoCo oracle} and the learned models, the errors increase for stiffer models.
In the case of learned models, the \textit{Hard} models perform worse when trained on up to $5000$ trajectories than the \textit{Soft} models perform only for $50$, a data-efficiency gap of at least $100$x.

\section{DISCUSSION}
\label{sec:discussion}
Our results provide compelling evidence that deep learning methods are negatively impacted by stiffness induced by contact.
Of particular note is that this effect is significant when compared to the inherent uncertainty of predicting stiff dynamics for noisy data; training set error grows nearly \textit{twice} as fast with stiffness as the \textit{MuJoCo oracle} model (Fig. \ref{subfig:train}).
We also see in Fig. \ref{subfig:generalization} that stiff dynamics can violate the common intuition that generalization error vanishes as the training set size approaches infinity.
While our softer models clearly behave as such, a $100$x increase in training set size made little impact on generalization for \textit{Hard} models.

It is also vital to understand how learning performance affects the downstream robotics task.
Often, a goal in robotics is to gain an \textit{accurate enough} model as fast as possible to use for long-term prediction.
The bottleneck in this scenario is often training data collection \cite{Nagabandi2019}.
From this data-efficiency perspective, increased stiffness has degraded learning performance 100-fold, as that the \textit{Hard} models are still worse than the \textit{Soft} models with 100 times more data (Fig. \ref{subfig:positionerror}--\ref{subfig:rotationerror}).

It is worth noting that long-term prediction performance can decorrelate from short-term predictions, especially for stiff dynamics \cite{Ajay2019,Halm2019}.
Training directly on multi-step prediction error is therefore desirable.
However, the stiffness of the training optimization problem grows \textit{exponentially} in the prediction horizon \cite{Ribeiro2019}, and we have seen here that even a prediction horizon of $1$ is challenging on stiff systems.
\section{CONCLUSION AND FUTURE WORK}
\label{sec:conclusion}
In this work, we have outlined a fundamental conflict between the dynamics of contact and common deep learning approaches, which significantly degrade the performance of learned models.
While compelling, these results are only an initial study.
Notably, our system has no actuation, and has lower dimensional state than many robotics tasks; the performance gap between hard and soft contact could be even wider for more complex systems.
Future studies into more complex systems are vital to understanding this relationship.
Hardware experiments would also further relevance of our results to real-world robotics.
Data efficiency has also been a primary focus of several recent contributions to robotic learning. 
Some approaches have focused on training on long-term prediction \cite{Ajay2018,Ajay2019}; and locally-accurate, task-specific models \cite{Chua2018}.
While none of these methods attempt to handle the conflict between stiffness and deep learning directly, examining effects of stiffness on these algorithms would strengthen the relevance of our results to the state-of-the-art.

The ultimate goal of quantifying stiffness's challenges is to inspire algorithms which can overcome them.
One direction is to shift deep learning's inductive bias towards models with near-discontinuity instead of explicitly embedding physics priors.
Some methods for instance attempt to capture sensitivity via multi-modality \cite{Fazeli2017,Arruda2017}.
A key challenge of contact is the huge quantity of modes \cite{Pfrommer2020}, whereas application of these methods have been limited to only a few.
Another option is to explicitly combine the structure of contact with deep networks.
While it requires a detailed system model, \textit{Residual physics} methods circumvent limiting mechanical assumptions by learning a DNN which corrects an imperfect physical model \cite{Ajay2018}.
While such methods have been shown to be data efficient, their ability to directly circumvent issues due to numerical stiffness has not been proven.
For instance, if a discontinuity in the physical model is incorrectly located in state space, the residual model will need to cancel out the incorrect discontinuity \textit{and} learn the correct one.
By contrast, our previous work \cite{Pfrommer2020} implicitly captures discontinuity by embedding DNN-parameterized geometries and friction into a physics-based simulator.
While this method is numerically well-behaved, it still requires significant knowledge of the physical properties of the system, and furthermore is restricted to inelastic impact and dry friction.
Future extensions of this work may alleviate the latter issue by combining residual learning with implicit discontinuity representations.

 \section*{ACKNOWLEDGEMENTS}
This work was supported by the National Science Foundation under Grant No. CMMI-1830218, an NSF Graduate Research Fellowship under Grant No. DGE-1845298, and a Google Faculty Research Award.

\appendices%

\section{1D Example}\label{adx:1DExample}
We now detail the prediction task explored in Figure \ref{fig:1DExample}.
The 1D system has state $x = [z;\;\dot z]$ and continuous-time dynamics
\begin{equation}
	\ddot z = \begin{cases}
		-9.81 & z > 0\,,\\
		-kz -2\sqrt{k}\dot z -9.81 & z \leq 0\,.\\
	\end{cases}\label{eq:1DDynamics}
\end{equation}
For each stiffness setting, data $(\dot z_t, \dot z_{t+1})$ are selected with an initial velocity $\dot z_t \sim \mathcal U([-3,5])$, and simulated \eqref{eq:1DDynamics} with initial condition $[1;\;\dot z_t]$ for $1$ \si{\second} to generate the final velocity $\dot z_{t+1}$. We add gaussian noise $(\sigma^2 = .01)$ to both velocities.

For each stiffness, we train 100 models on different sets of 20 training and 20 validation datapoints selected from this distribution; one such training set is displayed in Figure \ref{subfig:1DModels}.
Each model is an MLP with input $\dot z_t$, two hidden layers of width 128, and output  $\dot z_{t+1}$.
Models are trained with MSE loss, and terminated with early stopping with a patience of 10 epochs.
The Adam optimizer is used, with learning rate and weight decay separately tuned for each stiffness by grid search on $\{\num{1e-2},\num{1e-3},\num{1e-4}\}$ and $\{\num{1e-2},\num{1e-4},\num{0}\}$, respectively, to minimize ground truth MSE. The average prediction of these models with a 1 std. dev. window are plotted in Figure \ref{subfig:1DModels}.
\section{Simulation Details}\label{adx:DataCollectionDetails}
Here, we provide additional details on the physics of MuJoCo and the data generation process.
\subsection{Interpenetration in MuJoCo}
Here we discussion interpenetration $r$ in MuJoCo.
During contact, $r = 0$ may be considered a constraint which must be stabilized for reliable simulation; the seminal approach of Baumgarte \cite{Baumgarte1972} is to enforce penetration to obey dynamics inspired by a spring-mass damper: $\ddot r_B = -(kr + b\dot r)$.
To scale this idea to efficient multibody simulation, MuJoCo computes a convex optimization-based approximation \cite{Todorov2014}:
\begin{equation}
\ddot r \approx (1-d(r)) \ddot r_s + d(r) \ddot r_B\,,
\end{equation}
where $\ddot r_s$ is the acceleration due to gravity and inertia (-9.81 for the point mass in Figure \ref{fig:1DExample}), and $d(r) \approx 1$ is a user-specified function.
A thorough description of this function is available in the MuJoCo documentation online \footnote{\url{http://www.mujoco.org/book/modeling.html}}.
\subsection{Data Generation}
To generate an initial state, we generate a perturbation $\Delta x_{0}$ around the nominal state $x_{0,ref}$, an initial condition taken from the ContactNets dataset \cite{Pfrommer2020}:
\begin{multline*}
		x_{0, ref} =
		[0.186,\,0.026,\,0.122,\,-0.525,\,0.394,\,-0.296,\\ -0.678,\,0.014,\,1.291,\,-0.212,\,1.463,\,-4.854,\,9.870]\,,
\end{multline*}
where the cube center of mass is $\sim 0.12$\si{\meter} above the ground, with initial downward velocity of $\sim 0.2$ \si{\meter\over\second}.
This perturbation consists of $\Delta p_0 \sim \mathcal U ([-0.1,0.1]^3)$ \si{\meter\over\second} ; $\Delta q_{0} \sim Q\left(\theta \frac{v}{||v||_2}\right)$, a body-axis rotation of angle $\theta \sim \mathcal U([-1,1])$ \si{\radian} and axis $v \sim \mathcal U([1,1]^3)$; $\Delta \dot p_0 \sim \mathcal U([-0.1,0.1]^{3})$ \si{\meter\over\second}; and $\Delta \omega_0 \sim \mathcal U([-0.1,0.1]^{3})$ \si{\radian\over\second}.
Since the average trajectory length for
the Hard setting is 80 time-steps ($\sim 540$\si{\milli\second}), we truncate the \textit{Medium} and
\textit{Soft} trajectories to 80 time-steps to make the amount of data
per trajectory equal across all settings.
After a trajectory is generated from this initial condition, we add two forms of noise to the configrations $[p_t;\; q_t]$.
First we add constant-in-time noise to represent cumulative sensor drift: $\Delta p \sim \mathcal U([1, 1]^3)$ \si{\milli\meter} and $\Delta q \sim Q\left(\theta \frac{v}{||v||_2}\right)$, with $\theta \sim \mathcal U([-1,1])$ \si{\deg} and axis $v \sim \mathcal U([1,1]^3)$.
Then, we add i.i.d. noise to each measurement to model the small inconsistencies between measurements in the same fashion: $\Delta p_t \sim \mathcal U([.01, .01]^3)$ \si{\milli\meter} and $\Delta q_t \sim Q\left(\theta \frac{v}{||v||_2}\right)$, with $\theta \sim \mathcal U([-.01,.01])$ \si{\deg} and axis $v \sim \mathcal U([1,1]^3)$.
Velocities are reconstructed from the noisy data by inverting the finite difference equations \eqref{eq:PositionIntegration}--\eqref{eq:OrientationIntegration}.
While measurements have very small relative position noise, finite differencing amplifies the velocity noise by $\frac{1}{\Delta t} \approx 160 \si{\per\second}$.

\section{Learning Details}\label{adx:LearningDetails}
\begin{figure*}[t]
\begin{minipage}{0.34\hsize}
\centering
\resizebox{\hsize}{!}{
\begin{tikzpicture}[item/.style={circle,draw,thick,align=center},
itemc/.style={item,on chain,join},scale=0.1]
 \node[item,fill=gray!10] (h0) {$0$};
 \node[below right= 1em and 1em of h0,item] (phi1) {$\phi_\theta$};
 \draw[very thick,-latex] (h0.south east) -- (phi1.north west);
 \node[above right= 1em and 1em of phi1,item,fill=gray!10] (h1) {$z_{t-h+1}$};
 \draw[very thick,-latex] (phi1.north east) -- (h1.south west);
 \node[below = 1em of phi1,item,fill=gray!10] (x1) {$x_{t-h+1}$};
 \draw[very thick,latex-] (phi1.south) -- (x1.north);
 
 \node[below right= 1em and 1em of h1,item] (phi2) {$\phi_\theta$};
 \draw[very thick,-latex] (h1.south east) -- (phi2.north west);
 \node[above right= 1em and 1em of phi2,scale=2,font=\bfseries] (h2) {\dots};
 \draw[very thick,-latex] (phi2.north east) -- (h2.south west);
 \node[below = 1em of phi2,item,fill=gray!10] (x2) {$x_{t-h+2}$};
 \draw[very thick,latex-] (phi2.south) -- (x2.north);
 
 \node[below right= 1em and 1em of h2,item] (phi3) {$\phi_\theta$};
 \draw[very thick,-latex] (h2.south east) -- (phi3.north west);
 \node[above right= 1em and 1em of phi3,item,fill=gray!10] (h3) {$z_t$};
 \draw[very thick,-latex] (phi3.north east) -- (h3.south west);
 \node[below = 1em of phi3,item,fill=gray!10] (x3) {$x_t$};
 \draw[very thick,latex-] (phi3.south) -- (x3.north);
 
 \node[below right= 1em and 1em of h3,item] (phidec) {$\phi_{\theta,dec}$};
 \draw[very thick,-latex] (h3.south east) -- (phidec.north west);
 \node[right = 1em of phidec,item,fill=gray!10] (xnext) {$x_{t+1}$};
 \draw[very thick,-latex] (phidec.east) -- (xnext.west);
\end{tikzpicture}
}
\caption{The structure of our RNN predictors. $\phi$ is a recurrent unit (GRU), while $\phi_{dec}$ is an MLP decoder.\label{fig:RNNStructure}}
\end{minipage}
\begin{minipage}{0.65\hsize}
\centering
	{\includegraphics[width=.49\hsize]{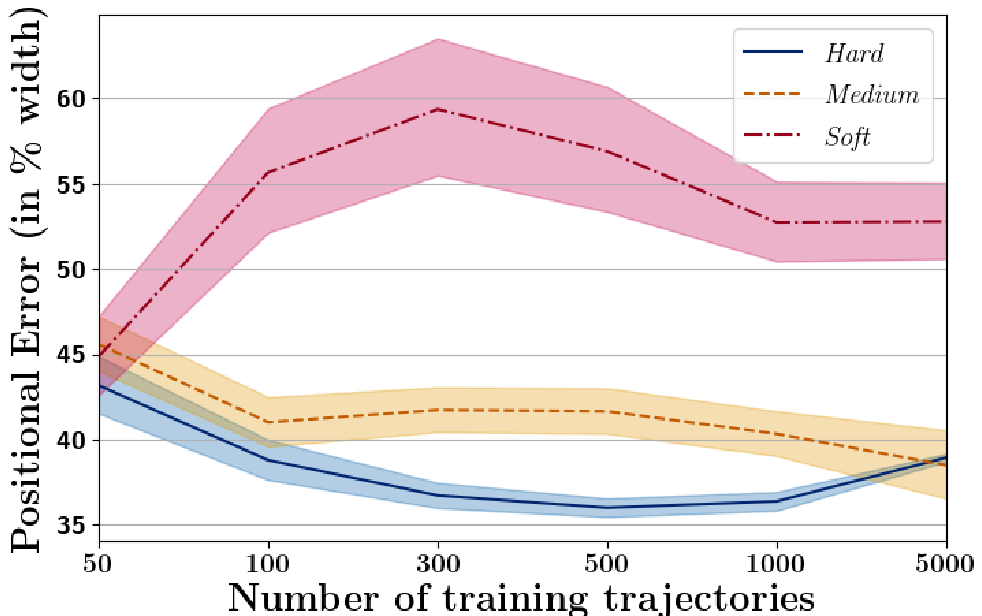}}
	{\includegraphics[width=.49\hsize]{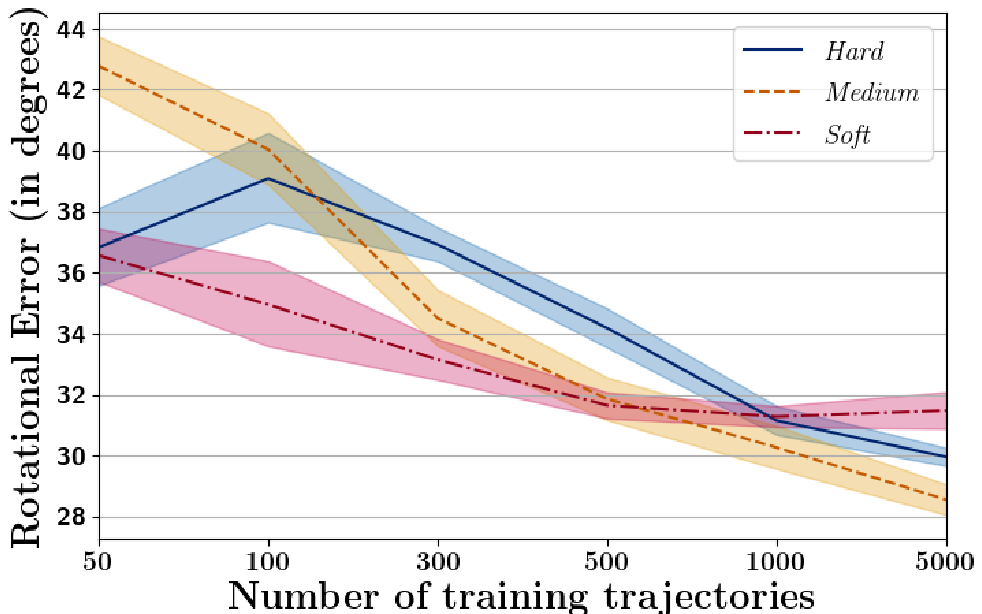}}
	\caption{We plot the performance of our optimized MLP networks on long-term prediction of position and orientation with $95$\% confidence intervals. \label{fig:MLPRolloutErrors}}
\end{minipage}	
\vspace{-5mm}
\end{figure*}

Our MLPs consist of 4 hidden fully-connected layers with ReLU activations, plus a final linear layer.
Our RNNs maintain a hidden state $z_t$ with initial value $0$. $z_t$ is updated sequentially for each $x_{t}$ as $z_{t} = \phi_\theta(x_{t}, z_{t-1})$ from the previous hidden-state $z_{t-1}$, where $\phi_\theta$ is a learned non-linear function.
By recursively unfolding (visualized in Fig. \ref{fig:RNNStructure}), 
\begin{equation}
	z_{t} = \phi_\theta(x_{t},\phi_\theta(x_{t-1},\dots\phi_\theta(x_{t-h+1},0) \dots ))\,.
\end{equation}
Finally, we use a two layer fully-connected network as a decoder $\phi_{\theta,dec}$ to extract the predicted velocity vector as $v_{t+1} = \phi_{\theta,dec}(z_t)$. The decoder consists of a hidden layer of width half the size of RNN hidden-state followed by ReLU activation units and the output layer. 
For each MLP and RNN architecture, we tried different target variables and sweep over different values of learning-rate, hidden-layer size, and weight-decay, centered around hand-tuned values.
While the history-length is $1$ for MLPs, for RNNs we also tried different history-lengths.
Table \ref{table:AllHyperparameters} provides the space of hyperparameters sweeped over in this process.
For each combination of settings, we complete at least $10$ training runs on high data regime of $500$ example trajectories, and then select the setting with the lowest average MSE over ${D_{\text{test}}}$.

\begin{table}
	\caption{Hyperparameter Search Space}
	\label{table:AllHyperparameters}
	\centering
	\begin{tabularx}{0.60\linewidth}{c c c c}
		\toprule
		Hyperparameter & \multicolumn{3}{c}{Values} \\
		
		\midrule
		target variable & $v_{t+1}$ & $\Delta{v}$ & -\\ 
		learning-rate & 1e-3 & 1e-4 & 1e-5 \\ 
		hidden-size & 128 & 256 & 512 \\
		history-length & 4 & 8 & 16 \\
		weight-decay & 0 & 4e-5 & 4e-3 \\
		\bottomrule
	\end{tabularx}
	\vspace{-6mm}
\end{table}

While MLP's and RNN's had similar training error and the generalization error trends, the long-term prediction error trends were noticeably different as demonstrated in Fig. \ref{fig:MLPRolloutErrors}. 
To make a fair comparison with our RNN's of history-length $h = 16$, our MLP rollout experiments also start from the $16$th time-step.
RNN’s are worse on \textit{Hard} models for rollouts, in spite of the fact that RNNs always perform better than MLPs on single-step predictions. 
We speculate that the learned models can exploit some sort of temporal consistency between velocities for smooth systems. 
After all, if the trajectory is smooth in time, then one can predict the velocity over a short horizon by simply doing polynomial extrapolation. 
However, as the velocity is nearly discontinuous in time for hard models, this intuition may no longer be valid.

\bibliographystyle{IEEEtran}
\bibliography{IEEEfull.bib,references.bib}

\end{document}